\documentclass[conference]{IEEEtran}
\IEEEoverridecommandlockouts
\usepackage{cite}
\usepackage{amsmath,amssymb,amsfonts}
\usepackage{algorithmic}
\usepackage{graphicx}
\usepackage{textcomp}
\usepackage{xcolor}
\usepackage{booktabs}
\usepackage{float}
\usepackage{multicol}
\usepackage{multirow}
\usepackage{afterpage}

\def\BibTeX{{\rm B\kern-.05em{\sc i\kern-.025em b}\kern-.08em
    T\kern-.1667em\lower.7ex\hbox{E}\kern-.125emX}}
\begin{document}

\title{LoRA-LiteE: A Computationally Efficient Framework for Chatbot Preference-Tuning\\
}

\author{\IEEEauthorblockN{Yahe Yang}
\IEEEauthorblockA{\textit{George Washington University}\\
yahe.yang@gwu.edu}

~\\
\and
\IEEEauthorblockN{Chunliang Tao}
\IEEEauthorblockA{\textit{New York University}\\
ct1942@nyu.edu}

~\\
\and
\IEEEauthorblockN{Xiaojing Fan}
\IEEEauthorblockA{\textit{New York University}\\
xf435@nyu.edu}

}

\maketitle

\begin{abstract}
Effective preference tuning is pivotal in aligning chatbot responses with human expectations, enhancing user satisfaction and engagement. Traditional approaches, notably Reinforcement Learning from Human Feedback (RLHF) as employed in advanced models like GPT-4, have demonstrated considerable success in this domain. However, RLHF methods are often computationally intensive and resource-demanding, limiting their scalability and accessibility for broader applications. To address these challenges, this study introduces LoRA-Lite Ensemble (LoRA-LiteE), an innovative framework that combines Supervised Fine-tuning (SFT) with Low-Rank Adaptation (LoRA) and Ensemble Learning techniques to effectively aggregate predictions of lightweight models, which aim to achieve a balance between the performance and computational cost. Utilizing the Chatbot Arena benchmark dataset, we conduct a comprehensive comparative analysis among our LoRA-LiteE model, corresponding base models at different scales, and GPT-4 trained with RLHF. Our empirical results demonstrate that the proposed LoRA-LiteE model achieves comparable performance to un-finetuned GPT-4 and outperforms the single larger-scale models under limited resource constraints. These findings highlight that our LoRA-LiteE provides a feasible and efficient methodology for human preference prediction in chatbot systems, enhancing scalability and accessibility, and thereby broadening the applicability of preference-tuned chatbots in resource-constrained environments.
\end{abstract}


\section{Introduction}
In recent years, chatbots have become integral to various industries, including customer service, healthcare, education, and entertainment \cite{adamopoulou2020overview, laranjo2018conversational, winkler2018unleashing}. These conversational agents serve as the first point of contact between organizations and users, providing information, resolving queries, and facilitating seamless interactions. The effectiveness of a chatbot is largely determined by its ability to generate responses that are not only contextually relevant but also aligned with human preferences and expectations \cite{gao2018neural}. This alignment, known as preference tuning, is pivotal in enhancing user satisfaction and engagement, thereby determining the overall success and adoption of chatbot technologies \cite{hancock2019learning, ziegler2019fine}.

Machine learning has been driving pivotal impacts in key sectors \cite{JIPD7671, Li_Wang_Chen_2024}. And traditional preference tuning methods have also employed various machine-learning techniques to align chatbot responses with human expectations. Early approaches, such as supervised fine-tuning and imitation learning, involved training models on large conversational datasets to replicate desired response patterns \cite{ross2011reduction}. While these methods have been effective, they often lack adaptability and a nuanced understanding of complex scenarios \cite{weston2016dialog}. In contrast, Reinforcement Learning from Human Feedback (RLHF) has emerged as the most advanced methodology for preference tuning. RLHF leverages feedback from human evaluators, enabling chatbots to iteratively refine their responses for better alignment with user preferences \cite{christiano2017deep, ouyang2022training}. This approach has been successfully implemented in state-of-the-art language models like OpenAI's GPT-4, which demonstrates exceptional capabilities in generating coherent, contextually appropriate, and preference-aligned responses \cite{achiam2023gpt}. The success of RLHF in models such as GPT-4 underscores its potential to create highly effective and user-centric chatbot systems.

Despite its successes, RLHF requires substantial computational and human resources, limiting the feasibility of deploying RLHF-based models in smaller organizations or for applications with limited budgets. To address these challenges, this study introduces LoRA-Lite Ensemble (LoRA-LiteE), an innovative framework that integrates Low-Rank Adaptation (LoRA) with ensemble learning to fine-tune lightweight chatbot models. Our approach offers several innovations: First, through parameter-efficient fine-tuning techniques such as LoRA, we significantly reduce the number of trainable parameters, enabling model training with limited computational resources \cite{hu2021lora}. Second, we adopt a multi-model ensemble strategy, leveraging the complementary advantages of different models to enhance overall system performance. Finally, we design targeted training strategies that optimize training efficiency while maintaining model performance. This integrated approach can not only alleviate the resource demands associated with RLHF but also improve the scalability and accessibility of preference-tuned chatbots, making advanced conversational agents feasible for a broader range of applications and organizations.

To evaluate the effectiveness of LoRA-LiteE, we utilize the Chatbot Arena benchmark dataset, a comprehensive collection of conversational interactions designed to assess chatbot performance across various tasks \cite{chiang2024chatbot}. Our experiments include LoRA-LiteE employing Llama-3-8b and Gemma-2-9b, single larger-scale LoRA-finetuned Llama-3-70b and Gemma-2-27b, and OpenAI's GPT-4 \cite{touvron2023llama, team2024gemma, achiam2023gpt}. Our results reveal that our LoRA-LiteE model outperforms the single fine-tuned llama-3-8b and gemma-2-9b models and achieves comparable accuracy to the un-finetuned GPT-4 in specific tasks, demonstrating LoRA-LiteE's capability to effectively align chatbot responses with human preferences. Notably, while the larger-scale models (Llama-3-70b and Gemma-2-27b) exhibit superior performance in environments with ample computational resources, their advantages diminish under limited resource constraints. As training time and available resources decrease, LoRA-LiteE begins to outperform these larger models, offering substantial computational efficiency and resource savings. These findings highlight the potential of our LoRA-LiteE architecture as an efficient and scalable alternative to RLHF methodologies, enhancing the accessibility and practicality of preference-tuned chatbots under resource constraints.

\section{Related Work}
\subsection{Preference Tuning}
Preference tuning, similar to efforts that improve LLM reasoning for greater interpretability and reliability, aims to align chatbot responses with human expectations, enhancing user engagement and overall experience in conversational systems \cite{christiano2017deep, zhang2024ratt, zhang2024thought}. The task is challenging due to the subjective and context-sensitive nature of human language, and existing approaches often require substantial computational resources and extensive annotated datasets. Reinforcement Learning from Human Feedback (RLHF) has emerged as a prominent method for aligning language models with human preferences by training a reward model from human feedback and fine-tuning the language model through reinforcement learning to maximize this reward \cite{ouyang2022training}. Models like GPT-4 have effectively utilized RLHF, yet the approach involves complex, multi-component training that requires considerable computational power and a large volume of human-labeled data, limiting its accessibility for smaller organizations. In contrast, Direct Preference Optimization (DPO), offers an alternative by directly optimizing the language model using preference data, eliminating the need for a separate reward model and complex reinforcement learning cycles \cite{rafailov2024direct}. Although DPO simplifies the training process and reduces some computational overhead, it still relies on large-scale models and considerable computational resources, which may not sufficiently address challenges faced in resource-constrained settings.

\subsection{Efficient LLMs}
Large Language Models (LLMs) have become indispensable across a range of tasks, from information extraction, question-answering systems to context learning, where they power complex tasks with superior performance \cite{lu2024context, zhao2024utilizing, yangusing, glenn2024blendsql, wang2023accurate}. Beyond language tasks, deep learning techniques, including reinforcement learning, have brought unprecedented improvements in applications like stock prediction, and autonomous vehicle navigation \cite{gu2024predicting, xu2024autonomousnavigationunmannedvehicle}. Despite their impressive performance, adapting these large models to specific applications often requires extensive fine-tuning and computational resources, making deployment costly and challenging. As such, many methods around efficient LLMs have been proposed to mitigate the issue \cite{xu2024can, yang2024cops}.

For efficient fine-tuning, Low-Rank Adaptation (LoRA) offers a distinct fine-tuning technique by freezing the original weights of the pre-trained model and injecting trainable low-rank decomposition matrices into each layer's weight matrices \cite{hu2021lora}. Specifically, it represents the weight updates as low-rank matrices added to the existing weights. By only updating these low-rank matrices, LoRA reduces the number of trainable parameters and lowers memory consumption during training. This enables fine-tuning large models on hardware with limited capacity without compromising performance, which is crucial for capturing the complexity necessary for effective preference tuning. LoRA is adopted in our work, as it provides a favorable trade-off between efficiency and performance and thus best aligns with our objective of developing preference-tuned chatbots that are both high-performing and feasible to deploy in environments with limited resources.

There are also extensive studies around model selections and integrations for efficient LLM deployment. For example, an SEO-LLM framework, combining model compression and data efficiency, has made significant advancements in deploying large-scale LLMs in under-resourced regions and industries \cite{li2024integrated}. Studies around selecting the appropriate LLMs that achieve the balance between performance and efficiency have provided feasible solutions in integrating LLMs into real-world applications under limited resources \cite{tao2024harnessing, xu2024restful, fan2024towards}. Federated Learning has also been explored to improve scalability, performance, and safety when data is limited and uncertain \cite{bonawitz2019towards, lin2021fednlp, zhang2022pmfl, zhang2024uncertainty}.

\subsection{Ensemble Learning}
Ensemble learning combines predictions from multiple models to improve overall performance, robustness, and generalization and has extensive applications in the Natural Language Processing (NLP) field \cite{liu2019roberta, clark2020electra}. By aggregating outputs from diverse models using techniques like bagging, boosting, and weighted voting, ensembles mitigate individual model biases and capture more nuanced relationships \cite{dietterich2000ensemble}. In machine translation, for example, ensembles of neural machine translation models have demonstrated significant improvements by leveraging the strengths of different models \cite{he2016dual}. Similarly, in stock price prediction, ensemble methods have demonstrated significant gains by leveraging the strength of different architectures to enhance predictive performance \cite{sui2024ensemble}. Additionally, combining models with different input formats, such as text, image, and audio, has proven to improve the robustness and safety in content generation \cite{tao2023meta, zhang2024steerdiff, tao2024nevlp}. Training multiple lightweight models and aggregating their predictions allows us to achieve high performance without the computational costs associated with large single models. This approach is particularly advantageous in resource-constrained environments, where deploying large-scale models is impractical due to limitations in computational resources or memory capacity. In our work, we leverage ensemble learning to combine multiple LoRA fine-tuned models, effectively balancing performance and computational efficiency to make preference-tuned chatbots more accessible and scalable.

\section{Methodology}

In this section, we present our approach to preference tuning for chatbots. Through ensemble learning, we leverage parameter-efficient fine-tuning methods like LoRA to achieve high performance while maintaining computational efficiency.

\subsection{Data Preprocessing}

The dataset provides 57,477 training samples, each containing a user query paired with responses from two different LLMs. We implemented a systematic preprocessing pipeline as detailed below.

\textbf{Dataset Structure } Each training sample includes four key components: \textit{Prompt}, \textit{Response A}, \textit{Response B}, and  \textit{Label}. \textit{Prompt} is user queries or instructions submitted to the models. \textit{Response A} and \textit{Response B} represents the generated output from Model A and Model B. \textit{Label} is the binary preference indicator denoting the winning model (Model A, Model B) or a tie condition.

\textbf{Data Cleaning} We perform several preprocessing steps to ensure data quality, including eliminating samples with null responses, transforming list-format strings into clean text, and structuring the input format.

\textbf{Tokenization Strategy}
We optimize the tokenization process by setting model-specific maximum sequence lengths, 1,536 tokens for Gemma-2 and 1,280 tokens for Llama-3, while implementing right-side padding with dynamic batch processing and adding EOS tokens.

\subsection{Training Framework}

\textbf{Training Mode Selection } During Supervised Fine-Tuning (SFT), we compare several training modes as shown in table \ref{table:training-modes}, from direct task-specific SFT to multi-stage training approaches. The best training mode for our preference prediction task is selected based on factors such as task requirements, dataset characteristics, and computational resources, with the goal to find the optimal balance between model performance and computational efficiency. 

\begin{table}[!htbp]  
\caption{Training Modes Considered}
\label{table:training-modes}
\centering
\renewcommand{\arraystretch}{1.3}
\begin{tabular}{ll}
\hline
\textbf{Mode} & \textbf{Configuration} \\
\hline
Mode 1 & Base model + Task-specific SFT \\
Mode 2 & Base model + Continue pre-train + Task-specific SFT \\
Mode 3 & Base model + Continue pre-train + General task SFT + \\
       & Task-specific SFT \\
Mode 4 & Base model + Continue pre-train + Mixed task and \\
       & domain SFT \\
Mode 5 & Base model + Continue pre-train + Mixed task SFT \\
Mode 6 & Chat model + Task-specific SFT \\
\hline
\end{tabular}
\end{table}
The selection criteria for the appropriate training mode are based on key factors, primarily the characteristics of task-specific datasets. For example, Mode 6 that combines chat model with task-specific SFT, is the best fit for our task. It is selected because targeted fine-tuning is far more efficient than continued pre-training given the modest-sized training and testing data. Additionally, accurately predicting human preferences between model responses typically requires robust dialogue understanding and evaluation capabilities, which aligns well with pre-trained chat models' strengths in conversational content comprehension.

\textbf{Fine-tuning } To optimize resource efficiency, we adopt LoRA (Low-Rank Adaptation) for parameter-efficient fine-tuning. This approach allows effective adaptation to the preference prediction task without updating all parameters, thereby reducing memory and computational requirements. Through targeted fine-tuning of a subset of model parameters, our system maintains high fidelity in predicting human preferences while preserving the model’s foundational capabilities in dialogue comprehension. LoRA's low-rank decomposition ensures that the model retains the semantic and contextual understanding essential for evaluating nuanced differences in response quality, all while achieving substantial reductions in computational overhead.

\textbf{Ensemble Strategy } Our approach incorporates an ensemble methodology at the end that combines predictions from multiple models using a weighted averaging strategy. This ensemble approach aims to leverage the complementary strengths of different models while maintaining robust prediction performance.

\section{Experiments \& Results}
\subsection{Experimental Setup}
In this study, we utilize the ChatBot Arena dataset, which contains 57,477 unique user interactions designed to evaluate and compare the performance of different Large Language Models (LLMs) \cite{chiang2024chatbot}. The dataset is collected through a human evaluation framework where judges interact with two different LLMs simultaneously, providing identical prompts to both models and selecting the more satisfactory response.

The dataset structure includes six key components: a unique identifier for each interaction, the identities of two competing models (A and B), the input prompt provided to both models, the corresponding responses generated by each model (from model A and B), and the judge's final decision. The judge's selection is encoded as a three-way classification, indicating whether model A, model B, or neither (tie) provided the more satisfactory response.

To assess the effectiveness of the proposed LoRA-Lite Ensemble (LoRA-LiteE) framework, we conduct a comparative analysis using the Chatbot Arena benchmark dataset. Our experiment encompasses six distinct models: two small base models, Gemma-2-9b and Llama-3-8b, which
are chosen for their lightweight architectures and proven efficiency in handling conversational tasks, facilitating effective ensemble integration; two larger base models, Gemma-2-27b and Llama-3-70b, chosen to provide a robust baseline for performance comparison; OpenAI's GPT-4, representing the state-of-the-art Reinforcement Learning from Human Feedback (RLHF) trained model; and our novel ensembled model, LoRA-LiteE, which integrates Gemma-2-9b and Llama-3-8b.

We fine-tune all models, except for GPT-4, on A100 GPU. The fine-tuning configurations are listed in table \ref{table:hyperparams}. The final ensemble prediction is calculated using empirically determined weights in Equation \ref{eq:ensemble}.
\renewcommand{\arraystretch}{1.3}
\begin{table}[!t]
\caption{Fine-Tuning Hyperparameters}
\label{table:hyperparams}
\centering
\begin{tabular}{lcc}
\hline
\textbf{Parameter} & \textbf{Gemma-2} & \textbf{Llama-3.1} \\
\hline
Learning Rate & $8 \times 10^{-5}$ & $1.2 \times 10^{-4}$ \\
Sequence Length & 1536 & 1280 \\
LoRA Rank & 32 & 32 \\
LoRA Alpha & 64 & 32 \\
Dropout & 0.0 & 0.0 \\
Frozen Layers & 2 & 2 \\
\hline
\end{tabular}
\end{table}

\begin{equation}
P_{final} = 0.7 \times P_{gemma} + 0.3 \times P_{llama}
\label{eq:ensemble}
\end{equation}

\subsection{Evaluation Metrics}

For each conversation in the test set, models must predict three probabilities that sum to 1 as shown in Equation (\ref{eq:model_output}).

\begin{equation}
    P(\text{winner\_model\_a}) + P(\text{winner\_model\_b}) + P(\text{winner\_tie}) = 1
    \label{eq:model_output}
\end{equation}

The performance of preference prediction models is evaluated using two metrics: log loss (also known as logarithmic loss or cross-entropy loss) and accuracy. Log loss, shown in Equation (\ref{eq:logloss}), is particularly suitable for assessing the quality of probabilistic predictions, while accuracy, shown in Equation (\ref{eq:acc}), provides an intuitive measure of model performance in multi-class classification problems. 

\begin{equation}
    \mathcal{L}(y, p) = -\frac{1}{N}\sum_{i=1}^{N}\sum_{j=1}^{M} y_{ij} \log(p_{ij})
    \label{eq:logloss}
\end{equation}

where:
\begin{itemize}
\item $N$ represents the total number of test samples
\item $M$ represents the number of classes (3 in our case: model\_a, model\_b, tie)
\item $y_{ij}$ is 1 if sample $i$ belongs to class $j$ and 0 otherwise
\item $p_{ij}$ is the predicted probability that sample $i$ belongs to class $j$
\end{itemize}

\begin{equation}
\text{Accuracy}(y, p) = \frac{1}{N}\sum_{i=1}^{N} \mathbb{1}(\arg\max_{j \in {a,b,tie}} p_j^i = \arg\max_{j \in {a,b,tie}} y_j^i)
\label{eq:acc}
\end{equation}
where:
\begin{itemize}
\item $N$ is the total number of test samples
\item $p_j^i$ is the predicted probability for class $j$ of sample $i$
\item $y_j^i$ is the ground truth label (0 or 1) for class $j$ of sample $i$
\item $\mathbb{1}(\cdot)$ is the indicator function
\item $\arg\max$ returns the class with highest probability/value
\end{itemize}

\subsection{Results and Discussion}
\renewcommand{\arraystretch}{1.3}
\begin{table}[H]
\centering
\caption{Models' Performances}
\label{table:best_performance}
\begin{tabular}{lcc}
\hline
\textbf{Model}  & \textbf{Accuracy(\%)} & \textbf{Log Loss}\\
\hline
Gemma-2-9b & 72.3 & 1.27 \\
Llama-3-8b	& 75.1 & 1.35\\
Gemma-2-27b & 84.8 & 0.72 \\
Llama-3-70b	& 86.9 & 0.79 \\
GPT-4 & 78.3 & 1.07\\
LoRA-LiteE & 80.2 & 0.99\\
\hline
\end{tabular}
\end{table}

\begin{figure}[!htb]
  \includegraphics[width=\linewidth]{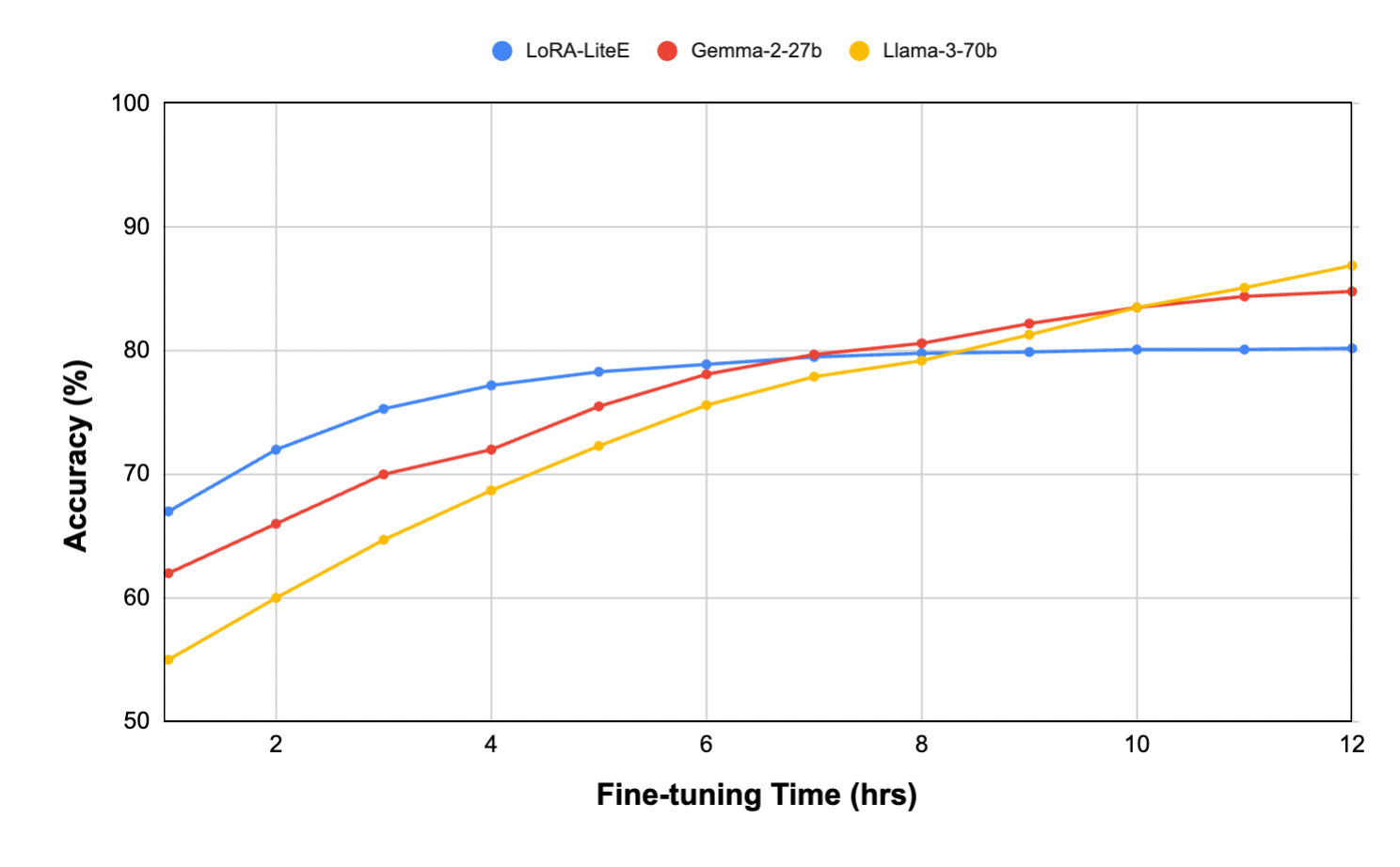}
  \caption{Comparative Accuracy Gains during Fine-tuning Across Models
}
  \label{fig:Comparative_Acc_Gain}
\end{figure}

\textbf{Best Performance Comparison } Table \ref{table:best_performance} summarizes the best performance achieved with early stopping criteria for each model. The larger-scale fine-tuned models demonstrated superior performances, attaining 84.8\% and 86.9\% accuracy, respectively. Our ensembled LoRA-LiteE model achieved the second-highest accuracy at 80.2\%, slightly outperforming GPT-4, which registered an accuracy of 78.3\%. The individual fine-tuned small base models, Gemma-2-9b and Llama-3-8b, exhibited lower accuracies of 72.3\% and 75.1\% respectively.

These results underscore the inherent performance advantages of larger models when ample computational resources are available. However, the ensembled LoRA-LiteE model demonstrates robust performance that rivals the advanced GPT-4, despite being based on smaller, lightweight architectures. Although GPT-4 is not fine-tuned for the specific task as other models do, it can still serve as a benchmark for RLHF-trained LLM. This indicates that ensemble techniques can effectively bridge the performance gap between smaller models and their larger counterparts, achieving competitive accuracy through the aggregation of predictions from multiple models.

\textbf{Efficiency vs. Accuracy Under Resource Constraints } Beyond absolute performance metrics, the efficiency of model fine-tuning is a critical factor, especially in resource-constrained environments. To evaluate the trade-off, we analyze the relationship between fine-tuning time (in hours) and test set accuracy of the LoRA-LiteE and the two larger counterparts, which have the best performance under convergence. Figure \ref{fig:Comparative_Acc_Gain} reveals that LoRA-LiteE achieves higher accuracy levels more rapidly compared to the larger-scale models Gemma-2-27b and Llama-3-70b within the initial seven hours of fine-tuning. The larger models continued to improve their accuracy beyond this time frame, ultimately attaining higher final scores.

The results demonstrate that as the available training time and computational resources decrease, the larger models' performance gains plateau and their ability to further enhance accuracy is significantly curtailed. In contrast, LoRA-LiteE not only maintains its competitive performance but also begins to outperform the larger models when fine-tuning time is restricted under certain threshold. This trend highlights the superior efficiency of LoRA-LiteE in scenarios where computational resources and time are limited, making it a more practical choice for organizations with constrained budgets or operational capabilities.

\textbf{Implications and Applications } The findings from our comparative analysis have significant implications for the deployment of preference-tuned chatbots in real-world settings. While larger models like Gemma-2-27b and Llama-3-70b offer higher accuracy in resource-rich environments, their substantial computational and memory requirements pose challenges to scalability and accessibility. LoRA-LiteE, with its ability to achieve comparable performance to GPT-4 and outperform larger models under resource constraints, presents a viable and efficient alternative. This advancement is particularly relevant for smaller organizations or applications operating under limited budgets, where deploying resource-intensive models is impractical. By leveraging lightweight architectures and ensemble learning, LoRA-LiteE facilitates the creation of sophisticated chatbot systems without the prohibitive costs associated with larger models. Furthermore, the reduced computational footprint of LoRA-LiteE contributes to lower energy consumption, addressing environmental concerns associated with extensive model training and deployment.

\section{Conclusion}
In this paper, we introduce LoRA-LiteE, a novel framework for preference optimization in large language models that combines parameter-efficient fine-tuning with ensemble learning. Our comparative performance analysis on the Chatbot Arena benchmark dataset shows that LoRA-LiteE not only surpasses individually fine-tuned smaller models but also rivals  much larger, resource-intensive models. This demonstrates LoRA-LiteE's ability to deliver high-quality results with reduced computational demands, expanding access to advanced AI tools for a broader range of developers and institutions across diverse applications. It also represents a significant step toward democratizing advanced AI technologies, fostering innovation, and facilitating the creation of AI systems that better serve the needs of diverse users.

While LoRA-LiteE demonstrates significant advantages in efficiency and scalability, it is essential to acknowledge certain limitations. Firstly, integrating multiple fine-tuned models into an ensemble adds complexity to the system. Managing and deploying multiple models can increase computational overhead during inference and complicate maintenance tasks like updates and debugging. This added complexity may limit the practicality of the approach in environments where simplicity and ease of deployment are crucial.  Future research could simplify the ensemble without sacrificing performance. One approach is model distillation, transferring the ensemble's knowledge into a single, smaller model to reduce complexity. In addition, our work focuses on general user preferences entailed by the dataset, and doesn't account for individual differences. This limits the effectiveness of the model in applications where customized responses are important for user satisfaction. Future work could incorporate personalization strategies to tailor responses to individual users. Lastly, the model assumes that user preferences are static, which doesn't reflect real-world scenarios where preferences evolve over time due to changing contexts or trends. This can result in the model becoming less effective as user expectations shift. Future work could focus on developing models that adapt to changing preferences. Implementing continuous learning mechanisms, such as online learning or real-time feedback loops, would allow the model to evolve with new data without retraining from scratch. Exploring time-aware modeling techniques can help the model anticipate and respond to shifting preferences, enhancing its long-term relevance and effectiveness.


\end{document}